\documentclass[review]{elsarticle}

\usepackage{amsmath}
\usepackage{mathtools}
\usepackage{adjustbox}
\usepackage{multirow}
\usepackage{textcomp}
\usepackage{lineno,hyperref}
\usepackage{multirow}
\usepackage{tabularx}
\usepackage{caption}
\usepackage{makecell}
\usepackage{subcaption}

\usepackage{lmodern}

\usepackage{colortbl}
\usepackage{pdflscape}
\usepackage{graphicx}
\usepackage{pgfplots}
\usepackage{longtable}
\usepackage{pgfplotstable}
\usepackage{cleveref}

\pgfplotstableset{
    /color cells/min/.initial=0,
    /color cells/max/.initial=1000,
    /color cells/textcolor/.initial=,
    color cells/.code={%
        \pgfqkeys{/color cells}{#1}%
        \pgfkeysalso{%
            postproc cell content/.code={%
                \begingroup
                \pgfkeysgetvalue{/pgfplots/table/@preprocessed cell content}\value
\ifx\value\empty
\endgroup
\else
                \pgfmathfloatparsenumber{\value}%
                \pgfmathfloattofixed{\pgfmathresult}%
                \let\value=\pgfmathresult
                \pgfplotscolormapaccess
                    [\pgfkeysvalueof{/color cells/min}:\pgfkeysvalueof{/color cells/max}]%
                    {\value}%
                    {\pgfkeysvalueof{/pgfplots/colormap name}}%
                \pgfkeysgetvalue{/pgfplots/table/@cell content}\typesetvalue
                \pgfkeysgetvalue{/color cells/textcolor}\textcolorvalue
                \toks0=\expandafter{\typesetvalue}%
                \xdef\temp{%
                    \noexpand\pgfkeysalso{%
                        @cell content={%
                            \noexpand\cellcolor[rgb]{\pgfmathresult}%
                            \noexpand\definecolor{mapped color}{rgb}{\pgfmathresult}%
                            \ifx\textcolorvalue\empty
                            \else
                                \noexpand\color{\textcolorvalue}%
                            \fi
                            \the\toks0 %
                        }%
                    }%
                }%
                \endgroup
                \temp
\fi
            }%
        }%
    }
}

\modulolinenumbers[5]

\begin{document}

\begin{frontmatter}

\title{An overview of artificial intelligence techniques for diagnosis of Schizophrenia based on magnetic resonance imaging modalities: Methods, challenges, and future works}

\author[mhd]{Delaram Sadeghi}
\author[labkntu]{Afshin Shoeibi\corref{mycorrespondingauthor}}
\cortext[mycorrespondingauthor]{Corresponding author}
\ead{afshin.shoeibi@gmail.com}
\author[labkntu]{Navid Ghassemi}
\author[mor]{Parisa Moridian}
\author[khd]{Ali Khadem}
\author[rol]{Roohallah Alizadehsani}
\author[khd]{Mohammad Teshnehlab}
\author[rem,remt]{Juan M. Gorriz}
\author[rol]{Fahime Khozeimeh}
\author[ukt]{Yu-Dong Zhang}
\author[rol,rolt]{Saeid Nahavandi}
\author[ach,acho,acht]{U Rajendra Acharya}

\address[mhd]{Department of Medical Engineering, Mashhad Branch, Islamic Azad University, Mashhad, Iran.}
\address[labkntu]{Faculty of Electrical Engineering, FPGA Lab, K. N. Toosi University of Technology, Tehran, Iran.}
\address[mor]{Faculty of Engineering, Science and Research Branch, Islamic Azad University, Tehran, Iran.}
\address[khd]{Department of Biomedical Engineering, Faculty of Electrical Engineering, K. N. Toosi University of Technology, Tehran, Iran.}
\address[rol]{Intelligent for Systems Research and Innovation (IISRI), Deakin University, Victoria 3217, Australia.}
\address[rem]{Department of Signal Theory, Networking and Communications, Universidad de Granada, Spain.}
\address[remt]{Department of Psychiatry. University of Cambridge, UK.}
\address[ukt]{Department of Informatics, University of Leicester, Leicester, UK.}
\address[rolt]{Harvard Paulson School of Engineering and Applied Sciences, Harvard University, Allston, MA 02134 USA.}
\address[ach]{Ngee Ann Polytechnic, Singapore 599489, Singapore.}
\address[acho]{Dept. of Biomedical Informatics and Medical Engineering, Asia University, Taichung, Taiwan.}
\address[acht]{Dept. of Biomedical Engineering, School of Science and Technology, Singapore University of Social Sciences, Singapore.}

\begin{abstract}
Schizophrenia (SZ) is a mental disorder that typically emerges in late adolescence or early adulthood. It reduces the life expectancy of patients by 15 years. Abnormal behavior, perception of emotions, social relationships, and reality perception are among its most significant symptoms. Past studies have revealed that SZ affects the temporal and anterior lobes of hippocampus regions of the brain. Also, increased volume of cerebrospinal fluid (CSF) and decreased volume of white and gray matter can be observed due to this disease. Magnetic resonance imaging (MRI) is the popular neuroimaging technique used to explore structural/functional brain abnormalities in SZ disorder, owing to its high spatial resolution. Various artificial intelligence (AI) techniques have been employed with advanced image/signal processing methods to accurately diagnose SZ. This paper presents a comprehensive overview of studies conducted on the automated diagnosis of SZ using MRI modalities. First, an AI-based computer aided-diagnosis system (CADS) for SZ diagnosis and its relevant sections are presented. Then, this section introduces the most important conventional machine learning (ML) and deep learning (DL) techniques in the diagnosis of diagnosing SZ. A comprehensive comparison is also made between ML and DL studies in the discussion section. In the following, the most important challenges in diagnosing SZ are addressed. Future works in diagnosing SZ using AI techniques and MRI modalities are recommended in another section. Results, conclusion, and research findings are also presented at the end.
\end{abstract}


\end{frontmatter}


\section{Introduction}
\begin{figure}[h]
    \centering
    \includegraphics[width=\textwidth ]{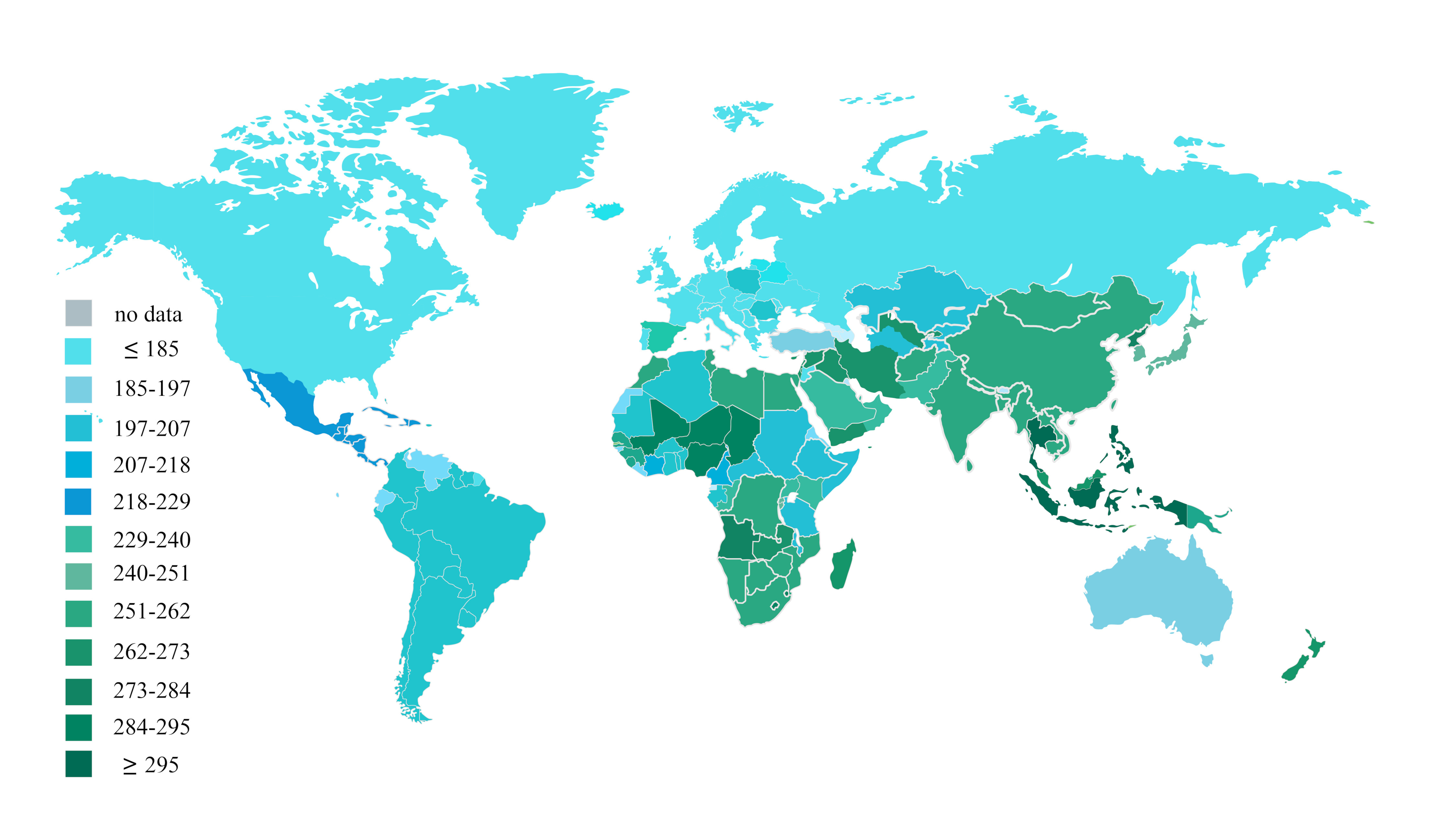}
    
    \caption{Regions showing the spread of SZ people around the world \cite{a7}}
    \label{fig:one}
\end{figure}

Schizophrenia (SZ) is the most severe psychological disease, which causes devastating effects on the brain and daily activities of the patient \cite{a1}. It causes abnormalities in the initial brain growth which may bring about different symptoms such as hallucination, disorder, motivational and cognitive problems \cite{a2}. The cause of this neural disorder is unknown, but neuroscientists believe that the interaction between genes and several environmental factors may be the main cause \cite{a2,a3}. Taking medicine reduces the psychological symptoms of SZ to some extent. However, these medicines do not improve the social and occupational activities of the patients completely \cite{a4}. According to the World Health Organization (WHO) reports, about 21 million individuals around the world suffer from this disorder. The average age for the onset of this disorder is 18 and 25 years in women and men, respectively with a higher prevalence rate in men \cite{a5,a6}. The regions showing the spread of SZ people around the world is illustrated in Fig. \ref{fig:one} \cite{a7}. 

Diagnosis of SZ is  a challenging problem due to the heterogeneity of this mental disorder and  lack of specific effective biomarkers \cite{a8}. In order to diagnose SZ, few clinical symptoms including physical, psychiatric, and psychological indicators need to be evaluated \cite{a9,a10,a11}. Clinical examination includes various tests such as blood tests as well as medical imaging \cite{a12,a13}. If the physicians do not find a physical cause for the suspected symptoms of SZ, they may refer the patient to a psychiatrist, psychologist, or other related experts. The main psychological assessment focuses on clinical interviews based on diagnostic and statistical manual (DSM-IV) of mental disorders conducted by clinical psychiatrists to diagnose patients with SZ \cite{a14,a15}.

Functional and structural neuroimaging techniques are another important category of methods capable to diagnose SZ \cite{a16,a17}. Structural neuroimaging modalities mainly include two methods of structural magnetic resonance imaging (sMRI) \cite{a18,a19,a20} and diffusion tensor imaging (DTI) \cite{a21,a22}, which show the structure of human brain and its structural connectivities, respectively, owing to their high spatial resolution. Overall, MRI based structural neuroimaging modalities are suitable for visualizing white matter (WM), gray matter (GM), and CSF tissues of the brain as well as exploring their abnormalities \cite{a23,a24}.

Functional neuroimaging modalities for the diagnosis of SZ include electroencephalography (EEG) \cite{a25}, magnetoencephalography (MEG) \cite{a26}, functional near-infrared spectroscopy (fNIRS) \cite{a27,a28}, and functional MRI (fMRI) \cite{a29,a30}. High cost and insufficient accuracy have limited the use of MEG and fNIRS for the diagnosis of SZ, respectively. 

The EEG is a noninvasive technique used to record the electrical activity of brain by using electrodes placed on the scalp \cite{a31,a32}. One of the problems with EEG is finding the exact location of brain activity sources \cite{a33,a34}.

The fMRI modality is one of the most studied techniques for diagnosing SZ and has two types of resting state (rs-fMRI) \cite{a35,a36} and task-based (T-fMRI) \cite{a37,a38}. The fMRI does not directly measure neural activity, but measures changes in blood oxygen, volume, and flow \cite{a35,a36,a37,a38}. During brain activities, regions of the brain involved in activity have higher blood flow than the rest, which increases oxygen levels \cite{a35,a36,a37,a38}. The better spatial resolution of fMRI over EEG and other functional modalities is one of the most important benefits which helps to determine nearly 1mm resolution where an activity occurs in the brain \cite{a39,a40}.

The limitations of sMRI and fMRI modalities are as follows. The common challenges of these two techniques the presence of noises and artifacts in the images. Hence, there is a need for stillness when recording the images to avoid high motion artifacts \cite{a41,a42}. Also, in fMRI, the temporal resolution is relatively low due to the slow hemodynamic response and also more time is needed to record a large volume of images \cite{a43,a44}. Hence, it is unable to monitor brain activities in real time \cite{a42,a43,a44}. These challenges make it difficult for physicians to accurately diagnose SZ.

Nowadays, computer aided diagnosis systems (CADS) have been proposed using advanced image processing and AI techniques to help the physicians to automatically diagnose SZ accurately \cite{a45,a46,a47}. Conventional machine learning (ML) and deep learning (DL) have been employed to develop highly accurate and robust CADS \cite{new1}. In this study, an extensive review is conducted on the diagnosis of SZ using functional and structural modalities of MRI and AI algorithms. 

The structure of this paper is as follows. Our search strategy is presented in Section 2. Then, in Section 3, the CADS for diagnosis of SZ based on MRI neuroimaging modalities is introduced and the related papers are reviewed. The main findings are discussed in Section 4. Subsequently, challenges in diagnosing SZ using AI techniques are discussed in Section 5. Finally, the paper is concluded and some future works are proposed in Section 6.

\section{Our Search Strategy }
\begin{figure}[h]
    \centering
    \includegraphics[width=0.85\textwidth ]{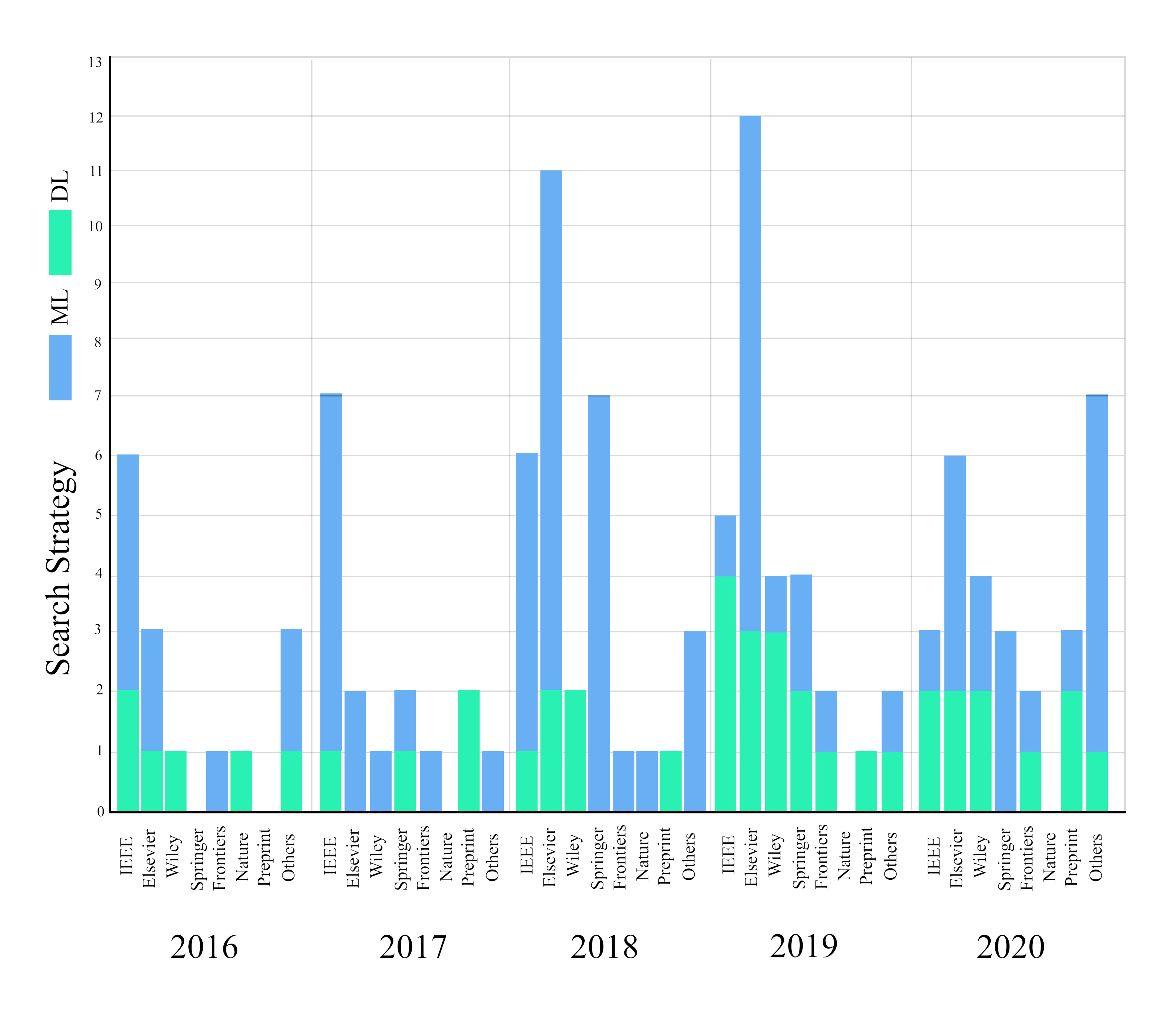}
    
    \caption{Number of papers published each year since 2016 on automated detection of SZ.}
    \label{fig:two}
\end{figure}

In this work, the most important citation databases such as IEEE Xplore, ScienceDirect, SpringerLink, and Wiley have been chosen to search articles on schizophrenia mental disorder. Also, the keywords "Schizophrenia", "sMRI", "fMRI", "Machine Learning", "Artificial Intelligence", and "Deep Learning" are employed in Google Scholar to find the relevant articles. The examination of the latest accepted papers until October 20\textsuperscript{th}, 2020 has been in this article. Figure \ref{fig:two} shows the number of papers published each year after 2016 on the automated detection of SZ using ML and DL techniques.  It can be noted from this figure that, Elsevier has published more number of papers as compared to the other publishers.

\section{CADS for Schizophrenia Diagnosis Based on Artificial Intelligence Methods}

\begin{figure}[h]
    \centering
    \includegraphics[width=\textwidth ]{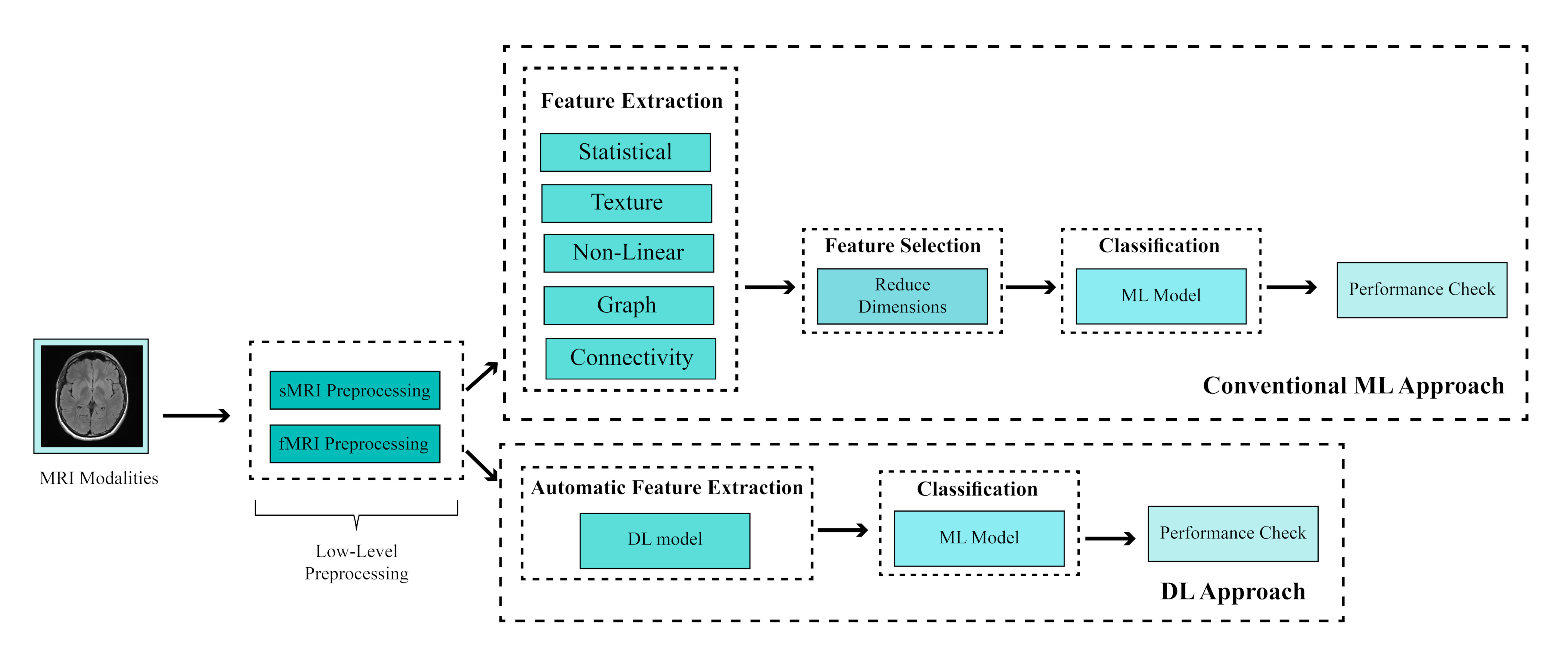}
    
    \caption{Illustration of automated diagnosis SZ using of AI techniques.}
    \label{fig:three}
\end{figure}

Nowadays, CADS is developed by the researchers to diagnose a variety of brain disorders such as epilepsy \cite{a48,a49}, autism spectrum disorders (ASD) \cite{a50,a51}, attention-deficit hyperactivity disorder (ADHD) \cite{a52,a53} and SZ \cite{a54,a55,a56} using MRI modalities. The implementation of CADS to diagnose SZ uses conventional ML or DL methods. These two categories of AI methods are graphically described in Figure \ref{fig:three}.

Figure \ref{fig:three} describes the ML methods for diagnosing SZ, in which the proper selection of feature extraction and feature selection methods requires extensive knowledge of image processing, feature engineering and AI. Also, the CADS steps for DL-based diagnosis of SZ are shown in Figure \ref{fig:three}. It can be seen that in DL, feature extraction and reduction/selection steps are merged into an automatic feature extraction step. Needing little knowledge of the field, intelligent and automatic representation learning, and good performance on big data are among the most important advantages of DL over ML. The important subsections of CADS for the automated diagnosis of SZ are presented in Figure \ref{fig:three}. 

\subsection{Available Datasets}
In this section, the freely available sMRI and fMRI neuroimaging datasets used for the diagnosis of SZ are introduced. Schizconnect \cite{a57}, NUSDAST \cite{a58}, COBRE \cite{a59}, FBIRN \cite{a60}, MCIC \cite{a61}, UCLA \cite{a62}, MLSP 2014 \cite{a63} are the important publicly available datasets for SZ detection. The details of these datasets are given below.

\subsubsection{Schizconnect}
This dataset has 1392 subjects used to diagnose SZ. In this dataset, 632 people have an undiagnosed disease, 215 people have broad SZ, 384 people have strict SZ, 41 people have schizoaffective disorder, 10 people have bipolar disorder, 44 people have sibling of SZ strict and 66 people have sibling of no known disorder \cite{a57}.
\subsubsection{NUSDAST}
This dataset can be downloaded as part of the SchizConnect dataset site. It contains various neuroimaging data obtained from 450 people with SZ, healthy controls, and their siblings over 2 years. Neuroimaging data includes sMRI, landmarks maps, FreeSurfer measurement, and segmentation. Cognitive data includes scores for crystallized intelligence, working memory, episodic memory, and executive performance. Clinical data includes demographics, sibling relationships, SAPS and SANS psychopathology. Genetic data is also 20 single nucleotide polymorphisms (SNPs). In addition to this dataset, CAWorks neuroimaging analysis software is also available. More information about this dataset is provided in \cite{a58}.
\subsubsection{COBRE}
A variety of neuroimaging modalities including rs-fMRI, sMRI, and phenotype and other diagnostic information from 72 patients with SZ and 75 healthy individuals (age range 18-65 years in each group) are included in this dataset. More information is provided in \cite{a59}.
\subsubsection{FBIRN}
It has three phases, with only second and third phases contain the data on people with SZ. Phase II consists of 87 individuals with DSM-IV SZ or schizoaffective disorder and 85 healthy individuals aged between 18-70 years. This dataset has sMRI of T1-weighted and T2-weighted contrasts. In addition, the third phase dataset includes neuroimaging modalities of DTI, sMRI, fMRI, and behavioral data, as well as clinical and demographic evaluations of 186 healthy individuals and 176 schizophrenics obtained from the United States \cite{a60}.
\subsubsection{MCIC}
This multi-site dataset contains a variety of sMRI, DTI, and fMRI neuroimaging modalities; all of which were obtained from 162 SZ patients and 169 healthy individuals. Clinical and cognitive assessments, genetic testing, etc. are also listed in this dataset. This dataset is available to the public through COINS \cite{a61}.
\subsubsection{UCLA}
This dataset contains various neuroimaging modalities including T-fMRI, rs-fMRI, sMRI, and diffusion-weighted imaging (DWI), as well as phenotype information obtained from 130 healthy individuals, 50 subjects with SZ, 49 subjects with bipolar disorder, and 43 subjects with ADHD. The description about the types of tasks for recording T-fMRI and preprocessing steps performed on images can be obtained from \cite{a62}.
\subsubsection{MLSP 2014 Schizophrenia Classification Challenge}
The MLSP dataset was introduced in a challenge held in 2014 under the auspices of the IEEE. This data contains sMRI and fMRI modalities recorded from 75 healthy individuals and 69 patients with SZ \cite{a63}. More detailed information about the dataset is given in Table \ref{tab:datasetOne}.

\begin{table}[h]
\centering
\caption{Details of freely available public  MRI datasets used for automated detection of SZ.}
    \label{tab:datasetOne}
    \resizebox{\textwidth}{!}{
    \begin{tabular}{|c|c|c|c|c|}
    \hline
    \textbf{Dataset} & \textbf{Publisher} & \textbf{Modalities} &\textbf{Number of Cases} & \textbf{Link}\\
    \hline
    SchizConnect & SchizConnect & sMRI, fMRI & \makecell{No Known Disorder=632\\Schizophrenia Broad=215\\Schizophrenia Strict=384\\Schizoaffective=41\\Bipolar Disorder=10\\Sibling of Schizophrenia Strict=44\\Sibling of No Known Disorder=66} & http://www.schizconnect.org/ \\
    \hline
    COBRE & -- & \makecell{rs-fMRI,\\Anatomical MRI} & Schizophrenia=72, Healthy=75 & \makecell{http://fcon\_1000.projects.nitrc.org/indi/retro/cobre.html\\https://data.mendeley.com/datasets/3h4mt7xryk/1}\\
    \hline
    NUSDAST & \makecell{NIH-Funded Data\\Sharing Project} & sMRI&\makecell{Schizophrenia=171, Healthy=170\\Non-Psychotic Siblings=44\\Healthy Siblings=66} & \makecell{https://central.xnat.org/app/action/DisplayItemAction/search\_value/\\NUDataSharing/search\_element/xnat:projectData/search\_field/xnat:\\projectData.ID}\\
    \hline
    \makecell{FBIRN Phase II\\ \hline FBIRN Phase III}& \makecell{The Function\\Biomedical\\Informatics\\Research Network} & fMRI & \makecell{Schizophrenia=87, Healthy=85 \\ \hline Schizophrenia=87, Healthy=85} & \makecell{https://www.nitrc.org/projects/fbirn/\\https://www.nitrc.org/projects/mcic/\\https://coins.trendscenter.org/}\\
    \hline
    MCIC & -- & \makecell{sMRI, fMRI,\\DWI} & Schizophrenia=162, Healthy=169 & https://openneuro.org/datasets/ds000115/versions/00001\\
    \hline
    UCLA & \makecell{UCLA Consortium\\for Neuropsychiatric\\Phenomics} & \makecell{fMRI, sMRI,\\DWI} & \makecell{Schizophrenia=50,\\Bipolar Disorder=49\\ADHD=43, Healthy=130} & https://openneuro.org/datasets/ds000030/versions/1.0.0\\
    \hline
    MLSP 2014 & IEEE & fMRI, sMRI & Schizophrenia=69, Healthy=75 & https://www.kaggle.com/c/mlsp-2014-mri\\
    \hline
    \end{tabular}}

\end{table}

\subsection{Preprocessing for sMRI and fMRI Modalities}

In this section, the most important low-level preprocessing techniques of sMRI and fMRI modalities are reviewed. The sMRI and fMRI neuroimaging modalities are usually very complex, difficult, and time consuming to analyze. In addition, one of the most important problems with MRI-based data is the presence of various artifacts which always pose a serious challenge to physicians and radiologists in accurately diagnosing the type of disease. Therefore, if appropriate methods are not used for preprocessing while analyzing MRI-based images, the diagnosis of brain diseases may be erroneous. To solve these problems, various software packages have been introduced in recent years to preprocess sMRI and fMRI modalities, the most important of which are FMRIB Software Library (FSL) \cite{a64}, brain extraction tools (BET) \cite{a65}, FreeSurfer \cite{a66} and SPM \cite{a67}. In the following sections, the important low-level preprocessing techniques for sMRI and fMRI neuroimaging modalities are discussed.

\subsubsection{Standard (Low-level) sMRI preprocessing steps}
Conventional preprocessing methods for sMRI images are introduced in this section. Important preprocessing steps for this data type include denoising, inhomogeneity correction, skull stripping, registration, intensity standardization, de-oblique, re-orientation, and segmentation \cite{a68,a69,a70}. They are briefly explained below.

\begin{enumerate}
    \item \textbf{Denoising:} The sMRI images are exposed to various noises during the recording process \cite{a68,a69,a70}. Classic filters \cite{a71,a72}, wavelet filters \cite{a73}, etc. are among the most common methods of noise removal in sMRI imaging \cite{a68,a69,a70}.
    \item \textbf{Inhomogeneity Correction:} The resulting defect in the coils of MRI scanner is seen as a low frequency change in signal intensity of sMRI images. Rectifying this artifact should be performed before any quantitative sMRI analysis \cite{a68,a69,a70}.
    \item \textbf{Skull-Stripping:} During sMRI imaging, brain and skull tissues are recorded. But the skull does not contain important information for processing. Therefore, when analyzing sMRI images, this part is removed by various methods \cite{a68,a69,a70}.
    \item \textbf{Registration:} In sMRI analysis, this stage of preprocessing is very common for merging different types of image modalities and sequences; also it can be used for transforming images into a common standard space such as MNI \cite{a68,a69,a70}.
    \item \textbf{Intensity standardization:} The sMRI obtained from different scanners will not have the same exact intensity, even if those scanners followed the same imaging protocol. Intensity standardization techniques attempt to correct these changes by a scanner-dependent manner \cite{a68,a69,a70}. Histogram matching techniques are the most commonly used technique for MRI intensity standardization \cite{a68,a69,a70}.
    \item \textbf{De-Oblique:}  During the sMRI recording process, the scan angle sometimes deviates from horizon to record the entire brain which is called the oblique scan. In such circumstances, data registration may be done with less noise, but it makes registration between different images difficult. Therefore, in some studies, de-oblique preprocessing is performed \cite{a68,a69,a70}.
    \item \textbf{Re-orientation:} It specifies the image orientation process settings. Differences in image orientation can lead to mis-registration. As such, images are transformed and re-oriented to have the same direction \cite{a68,a69,a70}.
    \item \textbf{Segmentation:} Segmentation of sMRI image divides it into different brain textures, including white matter, gray matter, and CSF or into distinct brain regions. Segmentation can be used for a variety of purposes. For example, segmentation for the normalization process or using a specific segmentation to generate a mask for a region of interest (ROI) \cite{a68,a69,a70}.
\end{enumerate}
\subsubsection{Standard (Low-level) fMRI preprocessing steps }

Important fMRI preprocessing techniques include removal of first N volumes, slice timing correction, motion correction and volume scrubbing, normalization, spatial smoothing, and temporal filtering \cite{a74,a75,a76}, each of which is described below.

\begin{enumerate}
    \item \textbf{Removal of the First N Volumes:} When a magnetic field is applied to the brain, the hydrogen spins orient themselves in the direction of the magnetic field, and it takes about 5 to 6 seconds for these spins to reach a steady state. Therefore, the volume images obtained in the first few seconds should be deleted to balance the signal and also let the patient get used to the device environment to reduce the artifacts of recorded fMRI data \cite{a74,a75,a76}.
    \item \textbf{Slice Timing Correction:} This step aims to make blood oxygenation level dependent (BOLD) time series of all voxels located in different slices to have the same reference time which is usually the acquisition time of the first slice \cite{a74,a75,a76}.
    \item \textbf{Motion Correction and Volume Scrubbing:} Motion correction is used to correct head movements during fMRI recording. Motion correction by aligning the data with a reference image tries to minimize the effect of movements on the data. This reference is usually the first volume. In the next step, an approach called volume scrubbing is performed, which means removing images that have very intense head movement artifacts \cite{a74,a75,a76}.
    \item \textbf{Normalization:} The size, shape, and anatomy of the brain vary from person to person, so inter-subject comparisons are necessary to allow images to be transferred to a standard template, or in other words to be spatially normalized. Currently, the most popular template is the MNI, however other templates are also available \cite{a74,a75,a76}.
    \item \textbf{Spatial Smoothing:} It involves a weighted averaging of BOLD signals of adjacent voxels. This process is persuasive on account of neighboring brain voxels being usually highly correlated in function and blood supply \cite{a74,a75,a76}.
    \item \textbf{Temporal Filtering:} In fMRI modality, the important information lies in the frequency band lower than 0.1 Hz. However, the components lower than 0.01 Hz are known to be slow drift of non-neural origin. Therefore, a band-pass filter with a frequency band of 0.009-0.08 Hz is usually used to remove undesired components \cite{a74,a75,a76}.   
\end{enumerate}

\subsection{Artificial Intelligence Methods}
As mentioned in the previous sections, AI methods include two important categories of ML and DL techniques. The AI methods are used to automatically detect SZ in this section. First conventional ML methods and then DL techniques are discussed.
\subsubsection{Conventional Machine Learning Methods}
The most important difference between CADS based on DL and conventional ML according to Figure \ref{fig:three} is in the blocks of feature extraction and feature selection. In this section, the most important steps of feature extraction and feature selection for the automated diagnosis of SZ are described in Table \ref{tab:related}. Tables \ref{tab:related} and \ref{tab:relatedtwo} show that diagnosis of SZ by conventional ML methods has been of more interest to researchers than by DL. The main reasons behind the popularity of traditional ML over DL are; (i) ML methods are still relatively common and widespread, (ii) works well with even a small dataset. The description of the CADS sections based on ML methods are given below.

\textbf{\textsc{Feature Extraction Techniques}}
Feature extraction is the most important part of ML technique based diagnosis of SZ. It can be noted from Table \ref{tab:related} that the most important feature extraction techniques employed for SZ detection using MRI modalities are statistical, textural, nonlinear, graph, and connectivity matrix.
\begin{enumerate}
    \item \textbf{Statistical:} Statistical moments are considered as the most basic feature extraction techniques, which include mean, variance, standard deviation, moments, and so on \cite{a77,new2}. Authors in \cite{a170,a179}, have used these methods to extract these features.
    \item \textbf{Texture:} Textural features are the important feature extraction technique used in medical images \cite{a78,a79}. Using these methods, important informations are extracted from the texture of images. Grey level co-occurrence matrices (GLCM)-based methods \cite{a80,a81}, and Gabor filters \cite{a82,a83} are the most important texture methods. Authors in \cite{a209,a214} have proposed a method to diagnose SZ based on textural features. 
    \item \textbf{Non-Linear:} Extracting nonlinear features from neuroimaging modalities is very useful and may increase the performance of SZ diagnosis \cite{a84}. In research \cite{a178}, authors used non-linear features for SZ detection. 
    \item \textbf{Graph:} Another group of features used for the diagnosis of SZ is based on graph models. In these methods, a graph is first constructed or extracted from the data in an innovative way. Then, with the help of graph and local graph properties, the data is displayed again. These methods can also be used to select an unsupervised feature \cite{a85,a86}. A number of studies have used graph-based features to diagnose SZ \cite{a149,a187}.
    \item \textbf{Connectivity Matrix:} Connectivity matrix feature extraction methods are the primary scheme of feature extraction used for processing DTI and fMRI neuroimaging modalities \cite{a87,a88}. These features provide an informative representation about the structure and function of brain. Functional connectivity matrix (FCM) \cite{a89,a90} and structural connectivity matrix (SCM) \cite{a91} are the methods used for fMRI and DTI modalities, respectively. In several works, the FCM technique are used to features from MRI images to detect SZ.

\end{enumerate}

\textbf{Feature Reduction / Selection Methods }

Choosing the right feature selection method when designing CADS improves the diagnosis performance of SZ. In addition, when the size of the data attribute space is very large, using an appropriate feature set helps to reduce the computational costs required to train the system. So far, several methods have been proposed for feature reduction or feature selection problems \cite{a92,a93,a94}. The important feature reduction and selection methods used in CADS systems for SZ diagnosis are discussed below.
\begin{itemize}
    \item \textbf{Feature Reduction Techniques:} In these methods, the feature matrix is first received and then is transferred from the input space to an output space of reduced dimension. In few studies, principal component analysis (PCA) technique has been used to reduce features and improve the specificity \cite{a95}.
    \item \textbf{Feature Selection Methods:} In these methods, an optimum subset of basic features is selected and used. Feature selection algorithms are divided into three types:  (i) supervised \cite{a96}, (ii) unsupervised \cite{a97}, and (iii) optimization \cite{a98}. They are briefly discussed below. 
    
\end{itemize}

\textbf{Supervised Feature Selection Methods} 

Methods for selecting supervised features include techniques based on Relief \cite{a99}, Fisher \cite{a100}, Chi-Squared \cite{a101}, and correlation \cite{a102} types. The details of these methods are given below.

\begin{enumerate}
    \item \textbf{Relief Feature Selection:} In this method, at each step, a sample is randomly selected from the samples in the dataset. Then, the degree of relevance of each attribute is updated based on the difference between the selected sample and two neighboring samples \cite{a99}. If one of the features of the selected sample differs from the similar feature in the neighboring samples of the same class, the score of this feature is reduced. On the other hand, if the same feature in the selected sample differs from the similar feature in the neighboring samples of the opposite class, the score of this feature increases \cite{a99}. Authors in \cite{a146}, have used relief algorithms  to select the features.
    \item \textbf{Fisher Feature Selection:} This technique selects attributes that minimize inter-class distances between samples, while maximizing the distance for intra-class ones; also, this method is often used for binary classification problems \cite{a48,a100}. Through this method, the importance (weight) of each feature is determined. In \cite{a184}, Fisher selection method is used to select the features. 
    \item \textbf{Chi-Squared Feature Selection:} Working based on the chi-square test \cite{a101}, this method tries to find features that have a relation with the input data (I.e., they are dependant). In order to be able to properly use this test to measure the relationship between various features in a data set and the target attribute, there must be two conditions for attribute attribution and independent sampling of attributes \cite{a101}. Authors in \cite{a148,a176} have used this  method to select the features in CADS for the diagnosis of SZ.
    \item \textbf{Correlation Based Feature Selection:} Correlation-based techniques are the supervised feature selection methods, few of such methods have also been used in CADS to diagnose SZ \cite{a102}. Correlation-based techniques have shown excellent performance for feature selection.
    
\end{enumerate}

\textbf{Unsupervised Feature Selection Methods}

The methods of variance,  mean absolute value of the differences, scatter ratio, Laplacein score, and finally the clustering are the important unsupervised feature selection methods used \cite{a103,a104,a105,a106}.  In \cite{a175}, agglomerative hierarchical clustering feature selection method is used to diagnose SZ. In another study, authors in \cite{a178} tested the fuzzy rough set method and achieved promising results.

\textbf{Feature Selection Based Optimization Methods}

It is the another class of feature selection techniques used in the diagnosis of SZ. Genetic algorithms (GA) \cite{a107,a108}, ant colony optimization (ACO) \cite{a109}, binary particle swarm optimization (BPSO) \cite{a110} and non-dominated sorting genetic algorithm II (NSGA-II) \cite{a111} have been used in various studies to select the features for the diagnosis of SZ.

\subsubsection{Deep Learning Methods}
DL is an emerging field which is widely used in neuroscience for the automated diagnosis of mental disorders such as bipolar disorder \cite{a112,a113}, personality disorders \cite{a114}, depression \cite{a115}, and schizophrenia \cite{a245}. In order to diagnose SZ using sfMRI and fMRI neuroimaging modalities, DL techniques have been used. As shown in Table \ref{tab:related}, most of the researches have focused on implementing various convolutional neural network (CNN) models to diagnose SZ. The reason for this choice is the excellent performance of CNNs using 2D and 3D data \cite{a116,a117,a118}. However, research has shown that these networks also performed very well on one-dimensional medical data. Autoencoders (AEs) \cite{a119,a120}, recurrent neural networks (RNNs) \cite{a119,a120}, deep belief networks (DBNs) \cite{a119,a120}, generative adversarial networks (GANs) \cite{a121}, CNN-AE \cite{a119,a120}, and CNN-RNN networks \cite{a119,a120} have also been used in few studies. They are briefly explained in the following subsections.

\textbf{Convolutional Neural Networks (CNNs)}

CNNs have been used for automated diagnosis of SZ, which includes 1D-CNN, 2D-CNNs, Inception, GANs, CapsNet, and finally 3D-CNN. The details of these models are given below.

\begin{enumerate}
    \item \textbf{1D and 2D CNN:} The computer vision and image processing have drawn the attention of many researchers' since 1960s \cite{a119,a120,new3}. Nevertheless, given the high dimensionality of images, image processing tasks, such as classification and segmentation, have always been considered as difficult tasks. In 2012, AlexNet, which is a form of deep neural networks with 2D convolutional layers is able to reach high accuracies for image classification tasks \cite{a119,a120}. Since then, many other models have been presented, aiming to improve the performance of prior ones, such as VGG \cite{a122,new4}, GoogleNet \cite{a123}. Also, other variations of 2D-CNN have been created to make them suitable for other data types such as 1D-CNN, which is more suitable for electroencephalogram (EEG) \cite{a124}. Figure \ref{fig:four} shows a sample 2D-CNN architecture used for automated detection of SZ using MRI modalities.
    \begin{figure}[h]
    \centering
    \includegraphics[width=\textwidth ]{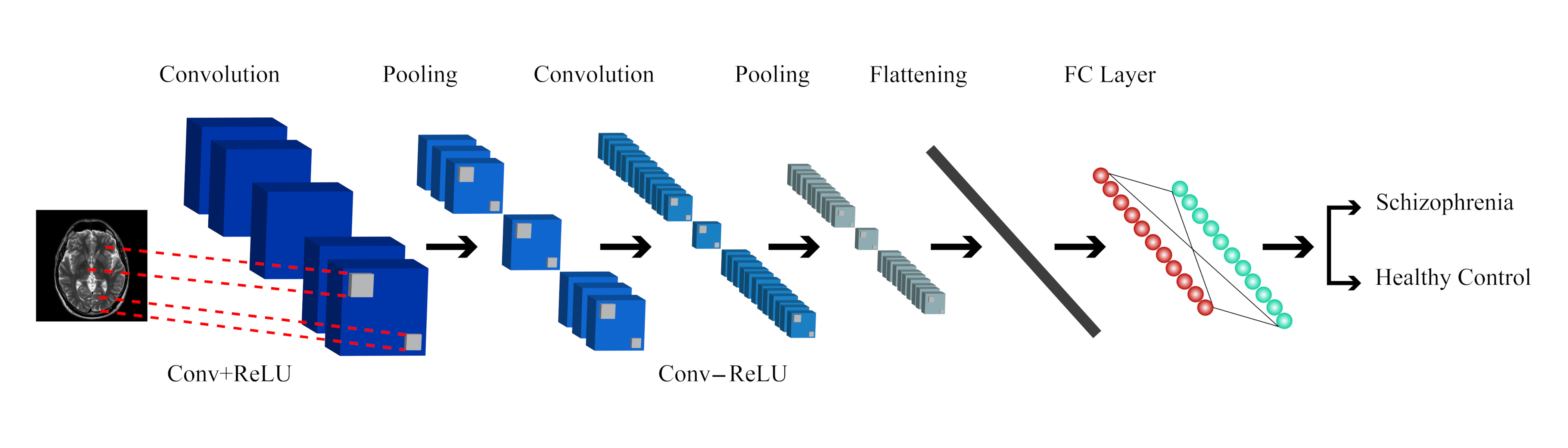}
    
    \caption{Block diagram of 2D-CNN used for automated SZ detection.}
    \label{fig:four}
    \end{figure}
    \item \textbf{Inception:} In the year 2014, two important network structures VGG and GoogLeNet are introduced. GoogLeNet, the winner of ImageNet challenge, had two primary ideas to overcome the vanishing gradient issue and go deeper \cite{a119,a120}. The first idea in this network is to use gradient injection, i.e., using a middle-level output for back-propagation additional to the last layer's output. The second and more important one is the inception layer. Inception layers combine the filters of various sizes to detect patterns of different lengths in the data. However, they also apply a 1x1 filter at the end of these blocks to reduce the number of parameters. Inception blocks are combined with many other structures to form more complicated and robust models, such as Inception-ResNet \cite{a125}.
    \item \textbf{Generative Adversarial Networks (GANs):} Generative models are not merely attractive due to their ability to generate new samples, but also the idea of making an algorithm that can generate samples itself is a significant step in creating intelligent models. However, the primary use of these models in biomedical data processing is to increase the size of datasets. Before GANs, many other generative models have been introduced. However, the quality of generated data samples is a concern in those models. Generative adversarial nets \cite{a121,a126} are first introduced mainly for images, and many other models have been  created for other data types also \cite{a121,a126}. In addition to generating new data, GANs can be used as unsupervised learning models as well \cite{a121,a126,new6}. Figure \ref{fig:five} shows a sample GAN architecture used for automated detection of SZ using MRI modalities.
    \begin{figure}[h]
    \centering
    \includegraphics[width=\textwidth ]{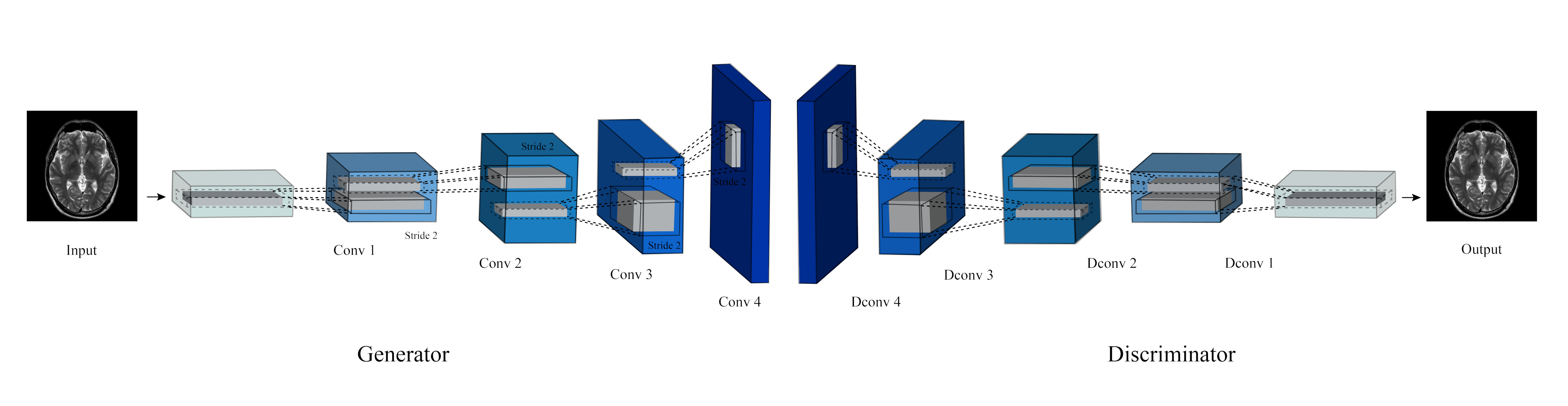}
    
    \caption{Block diagram of GAN used for automated detection of SZ.}
    \label{fig:five}
    \end{figure}
    \item \textbf{CapsNet:} The most important milestone in creating deep neural networks is to make them generalizable \cite{a119,a120}. Nowadays, many researchers try to do so by creating big datasets which contain various samples to include different situations that the data sample can be presented in; however, the CNNs' underperformance in the presence of data with a different orientation than training data stays as the primary deficiency of these models. CapsNet tried to address this issue by creating a network that implicitly performs reverse graphics \cite{a127,a128}. To achieve this, CapsNet proposed a block, capsule, which tries to determine the presence of an object in a given location and its instantiation. In recent years CapsNets have shown state-of-the-art performances in many applications \cite{a127,a128}. Figure \ref{fig:six} shows a sample CapsNet architecture used for automated detection of SZ using MRI modalities. 
    \begin{figure}[h]
    \centering
    \includegraphics[width=\textwidth ]{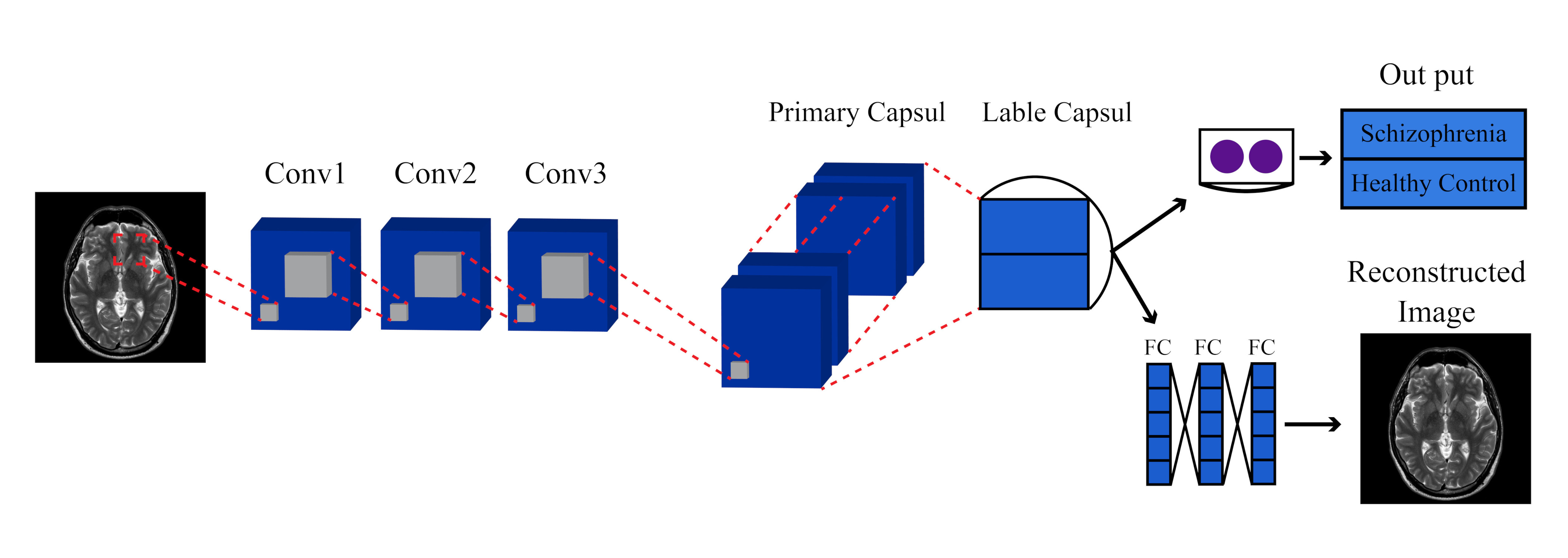}
    
    \caption{Block diagram of CapsNet used for automated detection of SZ.}
    \label{fig:six}
    \end{figure}
    \item \textbf{3D-CNN:} Convolutional neural nets perform well for 2D and 1D data due to their lower number of trainable parameters and transfer learning. However, for 3D datasets, designing and training a neural net is not as easy, considering the low volume of 3D datasets and a large number of trainable weights. Nevertheless, the possibility of finding spatial 3D patterns in the data has intrigued researchers to try to design and train 3D-CNNs despite their limitations \cite{a119,a120}. There are many 3D-CNNs have been developed to reach state-of-the-art performances \cite{a129,new5}. Figure \ref{fig:seven} illustrates a sample 3D-CNN architecture used for automated SZ using MRI modalities. 
    \begin{figure}[h]
    \centering
    \includegraphics[width=\textwidth ]{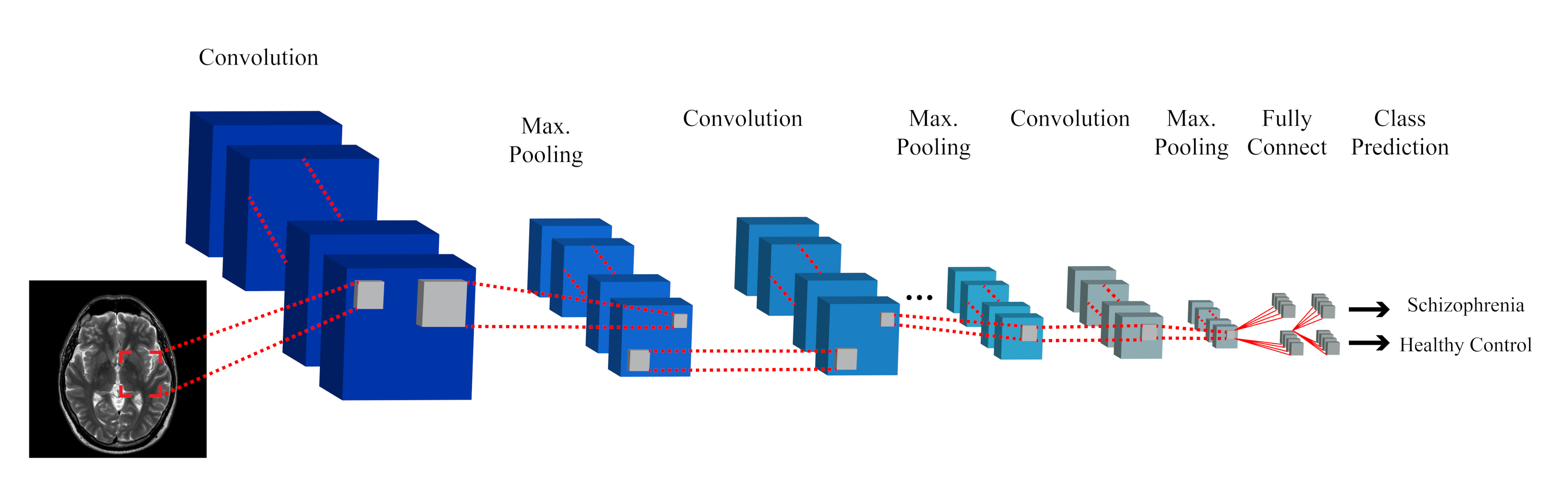}
    
    \caption{Block diagram of 3D-CNN used for automated SZ detection.}
    \label{fig:seven}
    \end{figure}
\end{enumerate}

\textbf{Recurrent Neural Networks (RNNs)}

Time-series and sequential data form a significant part of the data types. Recognizing temporal patterns while processing these data play a key role. However, the previous models developed are capable of recognizing spatial patterns, but they are not suitable for recognizing temporal patterns. Hence, recurrent neural nets (RNNs) are proposed to address this issue, which is a particular form of neural nets that can be scaled to detect distant patterns in time-series \cite{a119,a120}. Long short term memory (LSTM) and gated recurrent units (GRU) are the two famous building blocks of RNNs \cite{a119,a120}. 

\textbf{Autoencoders (AEs)}

\begin{figure}[h]
    \centering
    \includegraphics[width=\textwidth ]{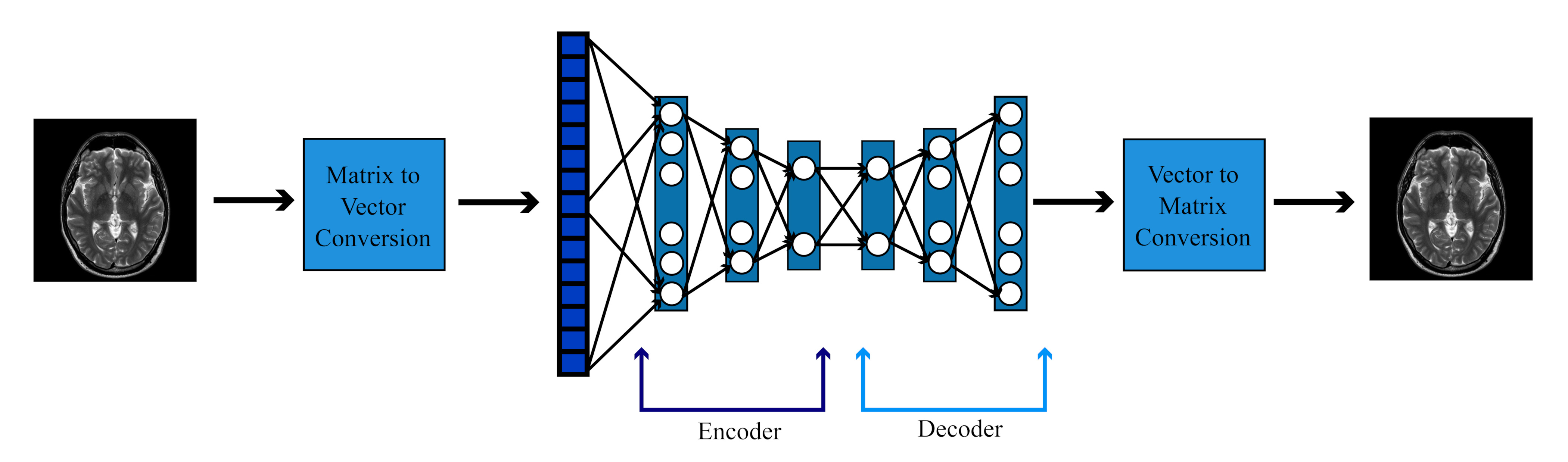}
    
    \caption{Block diagram of an AE used for automated SZ detection.}
    \label{fig:eight}
\end{figure}

Unsupervised learning is an exciting field in ML as it can eliminate all the overheads of feature engineering. Hence, AEs has been developed and used in many recent works \cite{a119,a120}. Basically, AEs try to map data to a smaller latent space by minimizing the loss function and then back to the original space \cite{a119,a120}. This moves the AEs toward preserving important characters of data while reducing its dimensionality. In recent years, many variants of AEs have been presented to improve their performance, such as stacked AE \cite{a130}, denoising AE \cite{a131}, and sparse AE \cite{a132}. Figure \ref{fig:eight} illustrates a sample AE architecture used for automated detection of SZ using MRI modalities.

\textbf{Deep Belief Networks (DBNs) }

Deep belief networks are a group of generative models created based on graphical models. These networks are composed of multiple layers of latent variables, and they have connections between layers but not within layers themselves. While they are considered as one of the premiers of the new era of DL and have been around for more than a decade \cite{a119,a120}, nevertheless, they are still widely used in recent studies with state-of-the-art performances.

\textbf{CNN-AE}

\begin{figure}[h]
    \centering
    \includegraphics[width=\textwidth ]{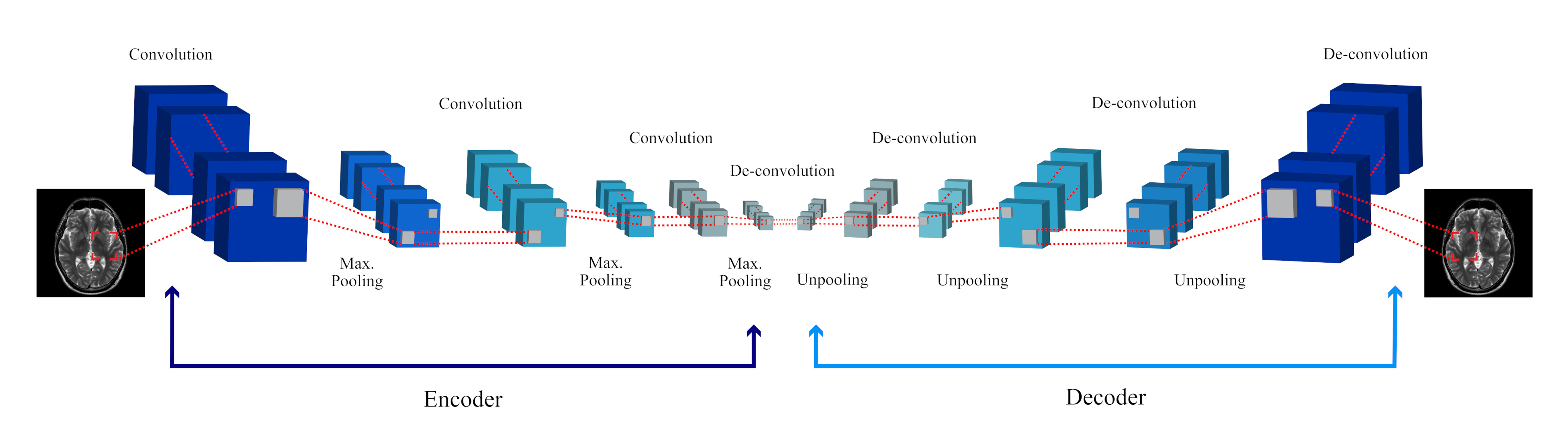}
    
    \caption{Block diagram of CNN-AE used for automated SZ detection.}
    \label{fig:nine}
\end{figure}

In order to use the benefits of convolutional layers in AEs for unsupervised representation learning, convolutional AEs (CNN-AEs), are introduced \cite{a48}. Figure \ref{fig:nine} illustrates a sample CNN-AE Architecture used for automated detection of SZ using MRI modalities. Due to the large number of learnable parameters, regular AEs usually overfit when fed with raw data, and they will not learn anything useful. So, applying convolutional layers can help to reduce the number of learnable parameters, and hence will make the network to adequately train. A combination of this model with others, such as sparse AE, can help to yield higher performance \cite{a119,a120,a48}. 

\textbf{CNN-RNN}

\begin{figure}[h]
    \centering
    \includegraphics[width=\textwidth ]{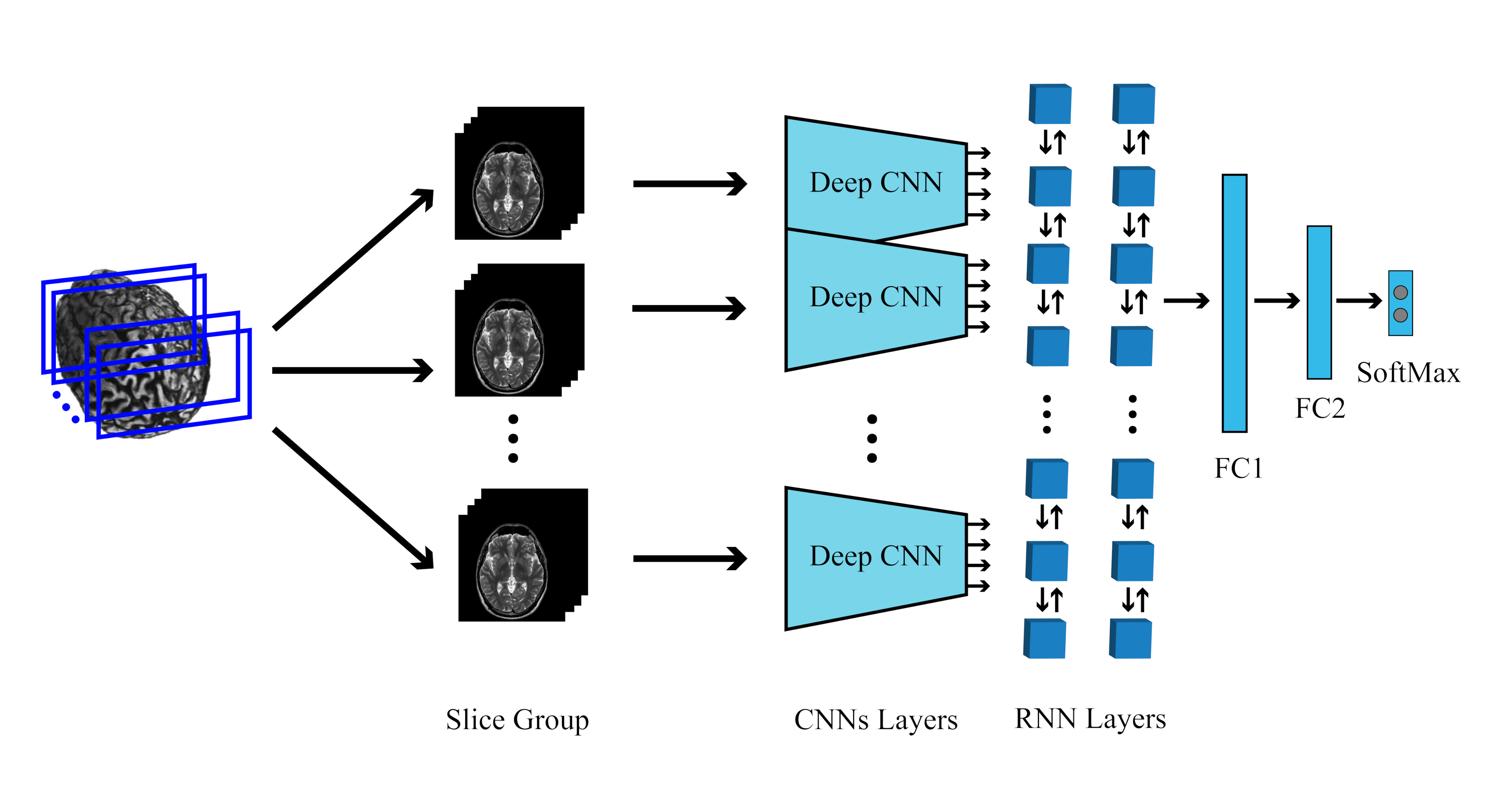}
    
    \caption{Block diagram of CNN-RNN used for automated SZ detection.}
    \label{fig:ten}
\end{figure}

While RNNs are strong in finding temporal patterns, they have issues when faced with spatial patterns \cite{a119,a120}. CNNs are the opposite, so if appropriately combined, a robust network capable of processing data with various types of characteristics, such as biomedical signals, can be created. Nowadays, CNN-RNNs are commonly used for signal processing tasks \cite{a133,a134}. In these networks, first few layers of convolution process data and extract features; then, these features are fed to RNN layers to make the final decision on the input \cite{a119,a120}. Figure \ref{fig:ten} illustrates the CNN-RNN network used for automated SZ detection. In this figure, many improvements can be made to this network, such as feature fusion to obtain higher performances \cite{a119,a120}. 

\subsubsection{Classification Methods}
The classification is the last part of CADS used to automatically detect SZ in DL. The support vector machines (SVM) \cite{a135,a136}, random forest (RF) \cite{a137} and Softmax \cite{a138} are among the widely used classification methods in the diagnosis of schizophrenia. Among the mentioned methods, Softmax method is used only in DL applications. On the other hand, SVM and RF classification techniques are used in both types of CADS, but the technique of implementing these methods is different in DL implementations \cite{a119,a120}. Details of the CADS implementation based on DL and conventional ML for the diagnosis of SZ are presented in Tables \ref{tab:related} and \ref{tab:relatedtwo}.

\clearpage

\begin{center}
\tiny
\setlength\LTleft{-130pt}            
\setlength\LTright{0pt}           
\begin{longtable}{|c|c|c|c|c|c|c|c|c|c|c|}

\caption{\centering Summary of studies conducted on the diagnosis of SZ using conventional ML and MRI modalities}
    \label{tab:related}\\
    \hline
    Work & Dataset & Miodalites & Number of cases & \makecell{High Level\\Prepreprocessing} & \makecell{Preprocess-\\ing\\Toolbox} & Feature extraction & \makecell{Feature\\reduction/\\selection} & Classifier & \makecell{K-\\fold} &\makecell{Performance\\criteria(\%)}\\
    \hline
    \multicolumn{1}{|c|}{\multirow{4}{*}{\shortstack[1]{\cite{a139}}}} & \multicolumn{1}{c|}{\multirow{4}{*}{\shortstack[1]{Clinical}}} & \multicolumn{1}{c|}{\multirow{2}{*}{sMRI}} & \multicolumn{1}{c|}{\multirow{4}{*}{\shortstack[1]{11 SZ, 11 HC}}} & \multicolumn{1}{c|}{\multirow{4}{*}{\shortstack[1]{Segmentation}}} & \multicolumn{1}{c|}{\multirow{2}{*}{\shortstack[1]{FreeSurfer}}} & \multicolumn{1}{c|}{\multirow{4}{*}{\shortstack[1]{Amplitude of Low\\Frequency Fluctuation\\(ALFF), GM\\Volume}}} & \multicolumn{1}{c|}{\multirow{4}{*}{\shortstack[1]{Linear Kernel\\Canonical Corre-\\lation Analysis\\(L-kCCA)}}} & \multicolumn{1}{c|}{\multirow{4}{*}{\shortstack[1]{NA}}} & \multicolumn{1}{c|}{\multirow{4}{*}{\shortstack[1]{--}}} & \multicolumn{1}{c|}{\multirow{4}{*}{\shortstack[1]{NA}}}\\
    \multicolumn{1}{|c|}{} & \multicolumn{1}{c|}{} & \multicolumn{1}{c|}{} & \multicolumn{1}{c|}{} & \multicolumn{1}{c|}{} & \multicolumn{1}{c|}{} & \multicolumn{1}{c|}{} & \multicolumn{1}{c|}{} & \multicolumn{1}{c|}{} & \multicolumn{1}{c|}{} & \multicolumn{1}{c|}{}\\
    \cline{3-3}
    \cline{6-6}
    
    \multicolumn{1}{|c|}{} & \multicolumn{1}{c|}{} & \multicolumn{1}{c|}{\multirow{2}{*}{rs-fMRI}} & \multicolumn{1}{c|}{} & \multicolumn{1}{c|}{} & \multicolumn{1}{c|}{\multirow{2}{*}{FSL}} & \multicolumn{1}{c|}{} & \multicolumn{1}{c|}{} & \multicolumn{1}{c|}{} & \multicolumn{1}{c|}{} & \multicolumn{1}{c|}{}\\
\hline
\end{longtable}
\setlength\LTleft{-129pt}            
\setlength\LTright{0pt}  
\begin{longtable}{|c|c|c|c|c|c|c|c|c|c|c|}

\caption{Summary of studies conducted on the diagnosis of SZ using DL and MRI modalities}
    \label{tab:relatedtwo}\\
    \hline
    Work & Dataset & \makecell{Modali-\\ties} & Number of cases & \makecell{High Level\\Prepreprocessing} & \makecell{Preprocessing\\Toolbox} & DNN & DNN toolbox & Classifier & K- fold &\makecell{Performance\\criteria(\%)}\\
    \hline
    \multicolumn{1}{|c|}{\multirow{3}{*}{\shortstack[1]{\cite{a219}}}} & \multicolumn{1}{c|}{\multirow{2}{*}{\shortstack[1]{NUSDAST}}} & \multicolumn{1}{c|}{\multirow{3}{*}{\shortstack[1]{sMRI}}} & \multicolumn{1}{c|}{\multirow{2}{*}{\shortstack[1]{141 SZ, 134 HC}}} & \multicolumn{1}{c|}{\multirow{3}{*}{\shortstack[1]{Segmentation}}} & \multicolumn{1}{c|}{\multirow{3}{*}{\shortstack[1]{NA}}} & \multicolumn{1}{c|}{\multirow{3}{*}{\shortstack[1]{Inception-\\ResNet}}} & \multicolumn{1}{c|}{\multirow{3}{*}{\shortstack[1]{NA}}} & \multicolumn{1}{c|}{\multirow{3}{*}{\shortstack[1]{SVM}}} & \multicolumn{1}{c|}{\multirow{3}{*}{\shortstack[1]{5}}} & \multicolumn{1}{c|}{\multirow{3}{*}{\shortstack[1]{Acc=70.98\\Spec=63.16\\Sen=75}}}\\
    \multicolumn{1}{|c|}{} & \multicolumn{1}{c|}{} & \multicolumn{1}{c|}{} & \multicolumn{1}{c|}{} & \multicolumn{1}{c|}{} & \multicolumn{1}{c|}{} & \multicolumn{1}{c|}{} & \multicolumn{1}{c|}{} & \multicolumn{1}{c|}{} & \multicolumn{1}{c|}{} & \multicolumn{1}{c|}{}\\
    \cline{2-2}
    \cline{4-4}
    \multicolumn{1}{|c|}{} & \multicolumn{1}{c|}{IMH} & \multicolumn{1}{c|}{} & \multicolumn{1}{c|}{148 SZ, 76 HC} & \multicolumn{1}{c|}{} & \multicolumn{1}{c|}{} & \multicolumn{1}{c|}{} & \multicolumn{1}{c|}{} & \multicolumn{1}{c|}{} & \multicolumn{1}{c|}{} & \multicolumn{1}{c|}{}\\
    \hline
    \cite{a220}	&\makecell{FBIRN phase-II\\fMRI}	&T-fMRI	&46 SZ, 49 HC	&tCompCor Denoising	&FSL	&LSTM	&NA	&NA	&10	&\makecell{Test\\Performance=\\66.4}\\
\hline
\end{longtable}

\end{center}

\section{Discussion}

\begin{figure}[h]
    \centering
    \includegraphics[width=0.75\textwidth ]{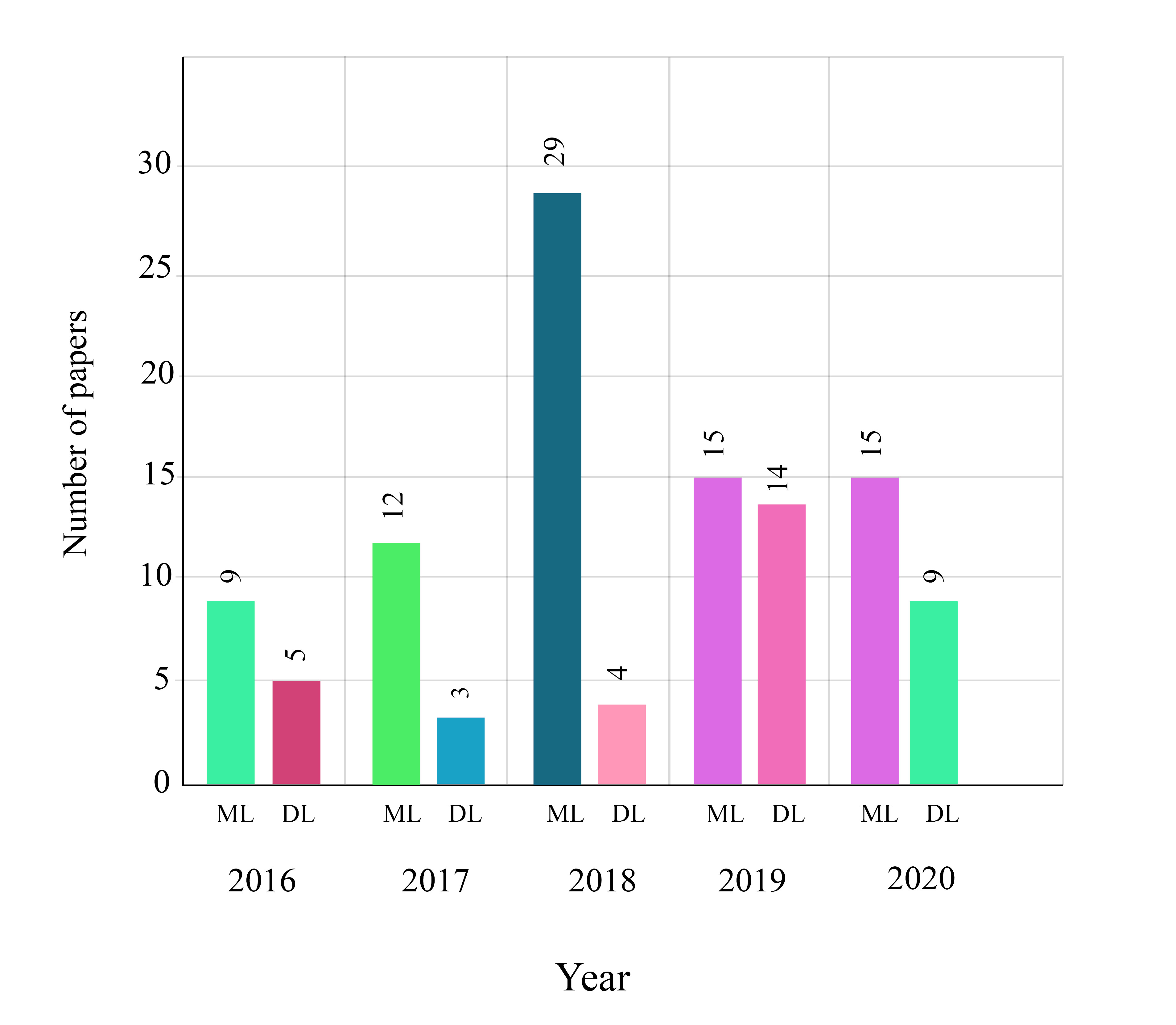}
    
    \caption{Number of papers published on conventional ML and DL for automated SZ detection using MRI modalities.}
    \label{fig:eleven}
\end{figure}

This paper provides a comprehensive overview of SZ diagnosis methods using MRI modalities and AI techniques. Since 2016 DL methods have paved the way for the diagnosis of SZ using MRI modalities; thus, only studies performed after that have been included in this review. The reason for this is to make a valid comparison between studies that used conventional ML methods over DL in the diagnosis of SZ. All studies on the diagnosis of SZ have been reviewed using sMRI and fMRI neuroimaging modalities along with DL and conventional ML methods in Tables \ref{tab:related} and \ref{tab:relatedtwo}. Table \ref{tab:related} shows the important information for schizophrenia diagnosis using conventional ML, including dataset types, modalities, preprocessing techniques, preprocessing toolboxes, feature extraction, feature selection / reduction, classification, K-Fold, and finally evaluation parameters. Similarly, Table \ref{tab:relatedtwo} focuses on DL methods, including DL architecture, DL toolboxes, and classification methods. Figure \ref{fig:eleven} shows the number of papers published on conventional ML and DL for automated SZ detection using MRI modalities.

It can be noted from Figure \ref{fig:eleven} that, conventional ML has been used more than DL for the automated SZ detection. This may be because there are limited numbers of public MRI datasets available. Secondly, conventional ML methods do not require powerful hardware resources, and by selecting less complex features, high performance can be achieved.

\begin{figure}[h]
    \centering
    \includegraphics[width=0.75\textwidth ]{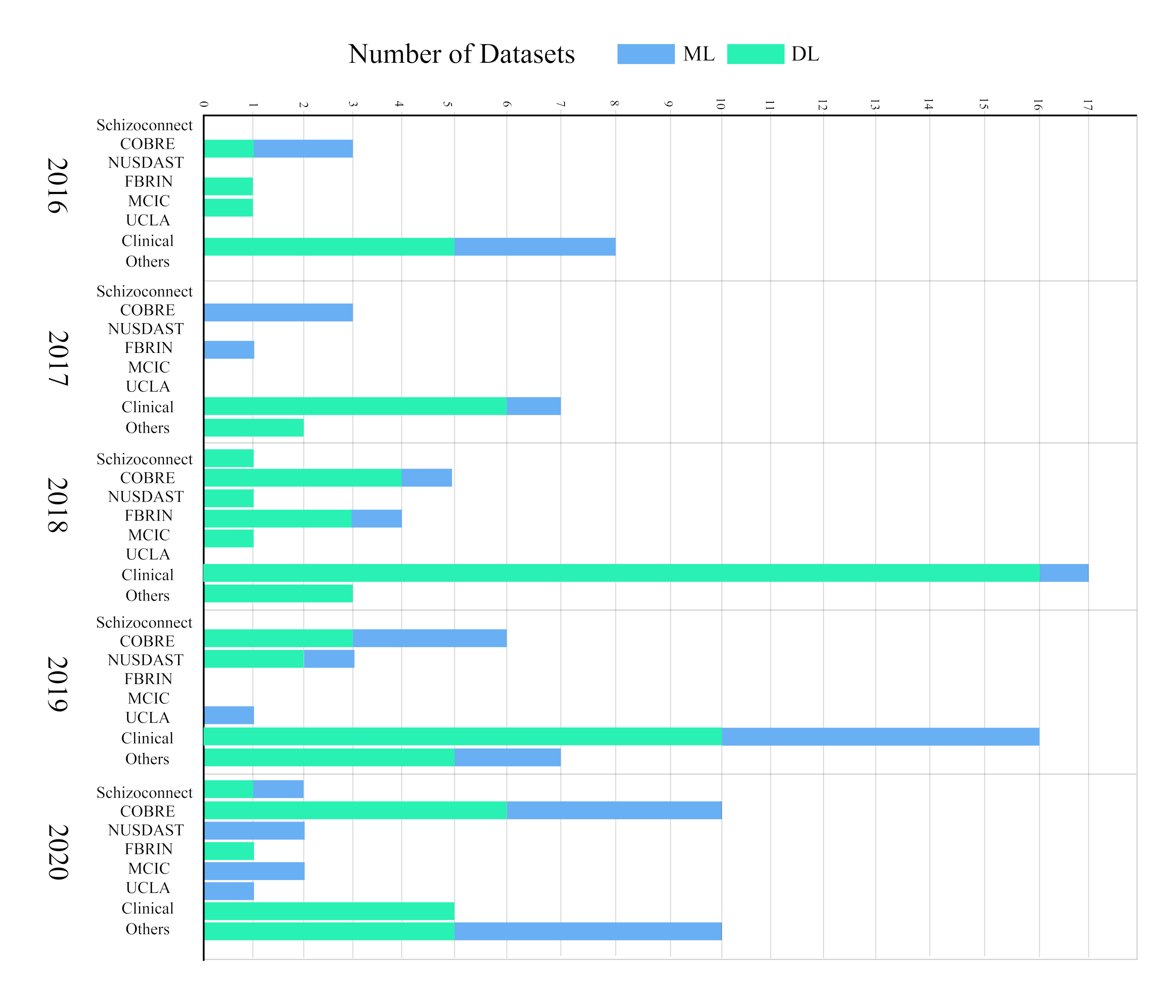}
    
    \caption{Number of studies published in the last four years on automated SZ detection using different MRI-based datasets.}
    \label{fig:twelve}
\end{figure}

It can be noted from Table \ref{tab:datasetOne} that, several freely available datasets are available for automated diagnosis of SZ. The various methods which have been proposed using these free datasets are shown in Tables \ref{tab:related} and \ref{tab:relatedtwo}. The number of datasets used to develop DL and ML models proposed each year is displayed in Figure \ref{fig:twelve}. It can be noted from this figure that, the COBRE dataset is more efficient and popular than other datasets for the studies on automated detection of SZ. As illustrated in Figure \ref{fig:twelve}, the COBRE dataset is of more significance than the rest because the size of normal and schizophrenia classes are equal in this dataset. 

\begin{figure}[h]
    \centering
    \includegraphics[width=0.75\textwidth ]{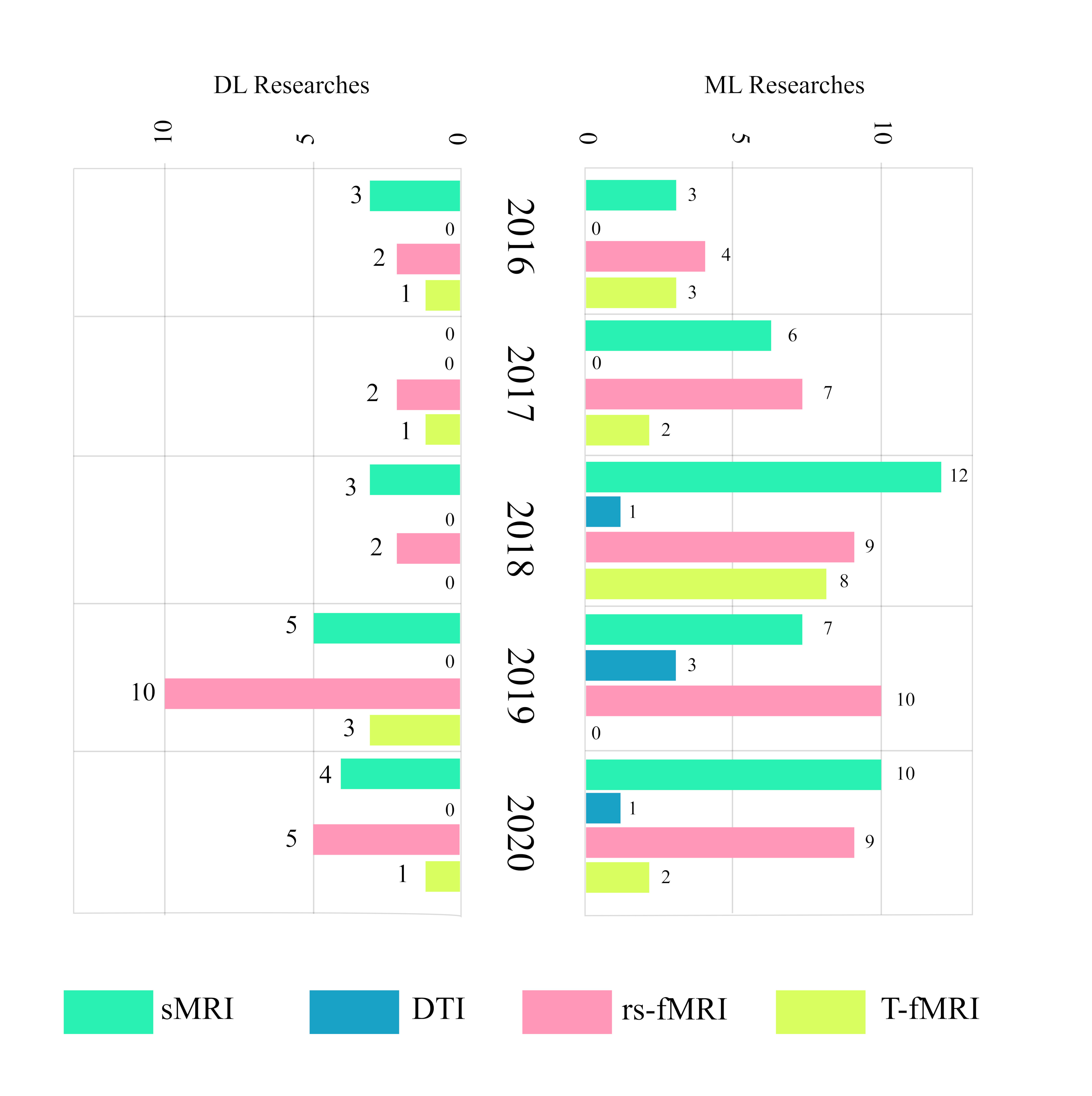}
    
    \caption{Number of studies published in the last four years on automated SZ detection using different MRI modalities and AI methods.}
    \label{fig:thirteen}
\end{figure}

Figure \ref{fig:thirteen} shows the types of sMRI and fMRI neuroimaging modalities used for the diagnosis of SZ in Tables \ref{tab:related} and \ref{tab:relatedtwo}. It can be seen that rs-fMRI neuroimaging modality has been widely used. Also, it can be noted from the figure that, in recent years more studies have been conducted on automated SZ using sMRI and rs-fMRI neuroimaging modalities. 

\begin{figure}[h]
    \centering
    \includegraphics[width=0.6\textwidth ]{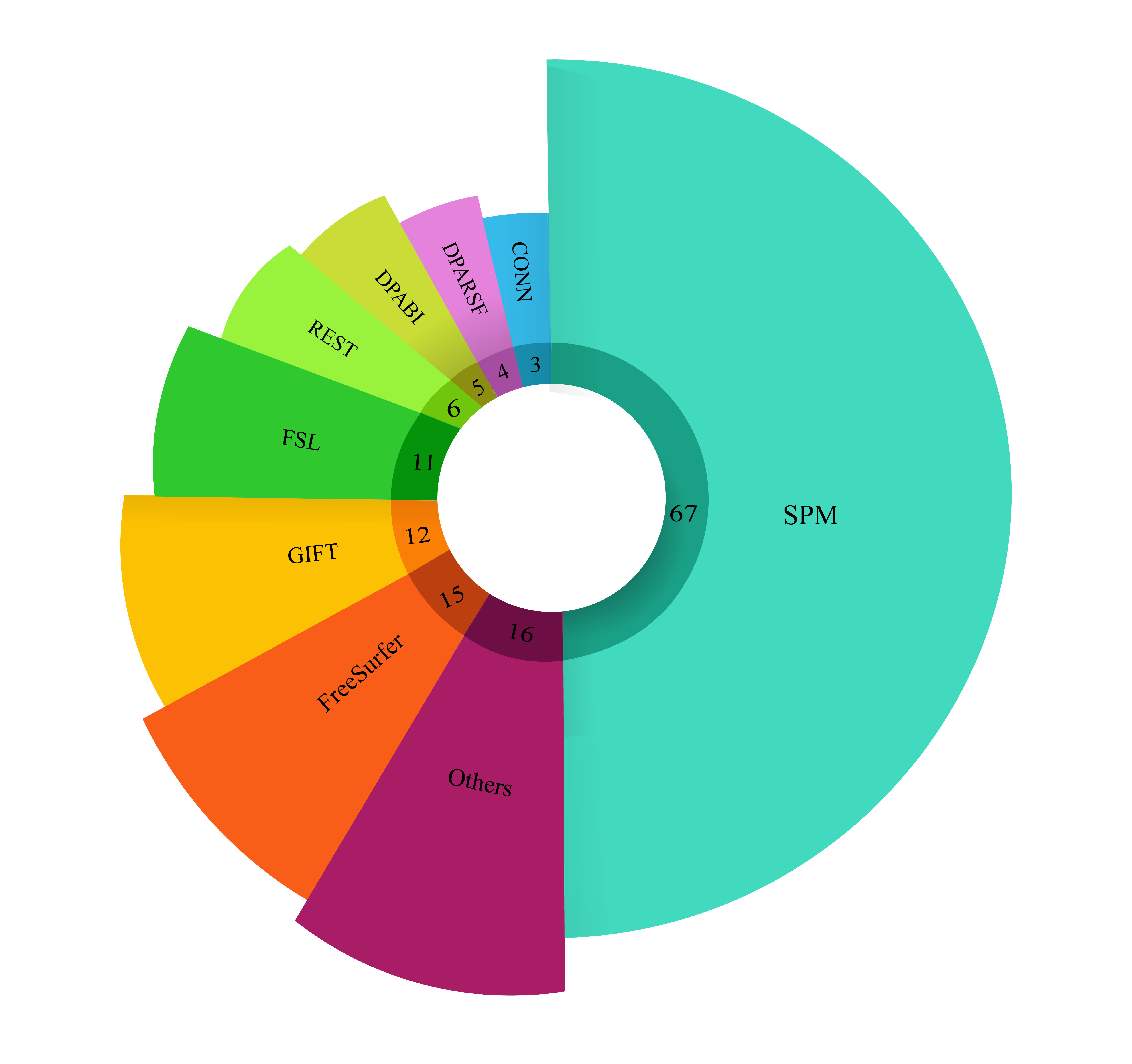}
    
    \caption{Number of preprocessing tools used for the automated diagnosis of SZ.}
    \label{fig:fourteen}
\end{figure}

Preprocessing of sMRI and fMRI modalities is an important step in the automated detection of SZ. The preprocessing techniques are divided into two categories of low-level and high-level methods, which were described in details in the previous sections. Low-level preprocessing using sMRI or fMRI modalities have specific and standard steps. Hence, FSL \cite{a64}, BET \cite{a253,a254}, FreeSurfer \cite{a66} and SPM \cite{a67} tools have been introduced for low-level preprocessing. The number of preprocessing tools used for the automated diagnosis of SZ is shown in Figure \ref{fig:fourteen}. It can be noted from this figure that, SPM toolbox has been widely used by the researchers. 

\begin{figure}[h]
\centering
\begin{subfigure}[b]{0.45\textwidth}
    \centering
    \includegraphics[width=\textwidth]{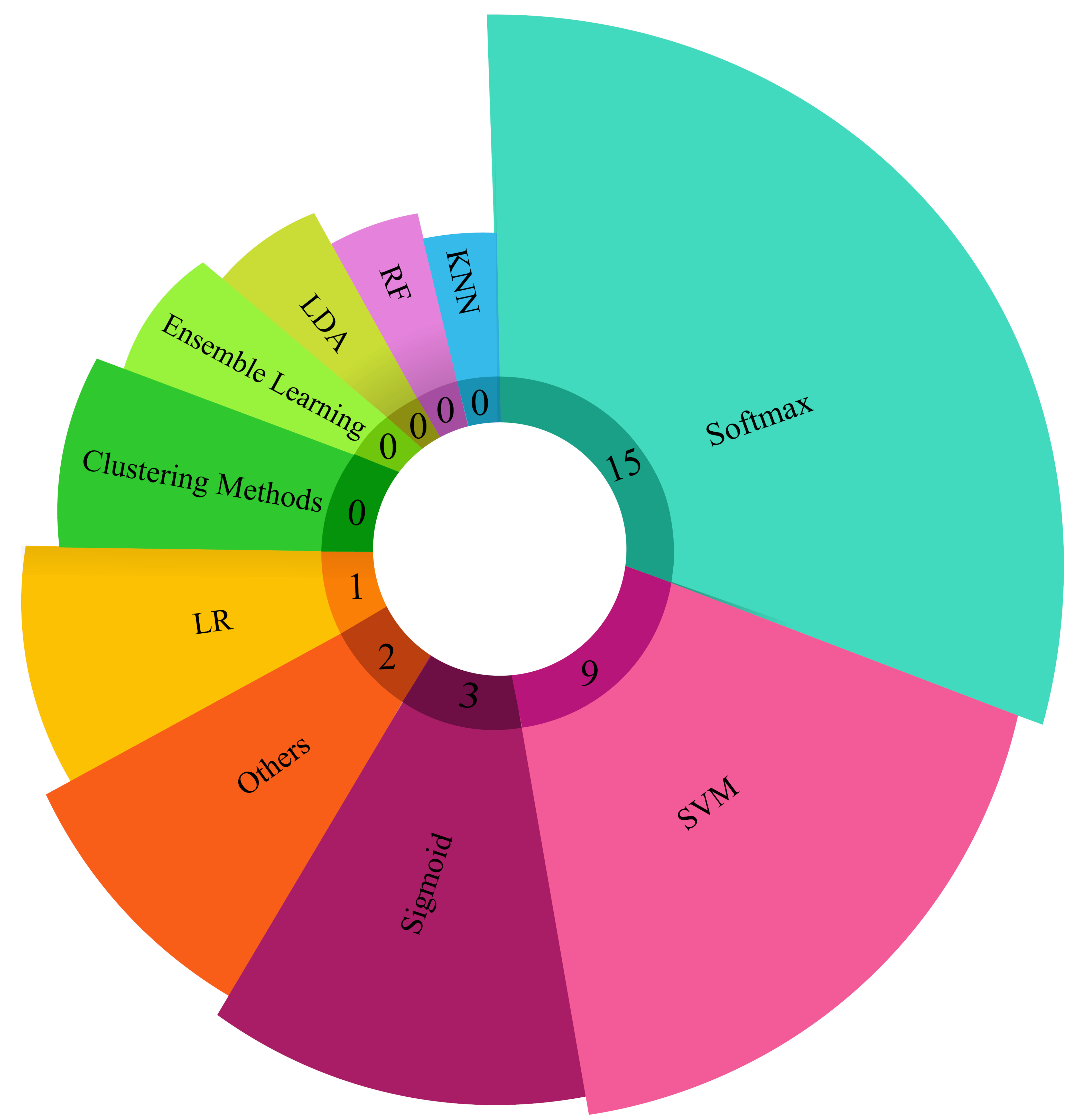}
    \caption{}
    \label{fig:15a}
\end{subfigure}
\hspace{10pt}
\begin{subfigure}[b]{0.45\textwidth}
    \centering
    \includegraphics[width=\textwidth]{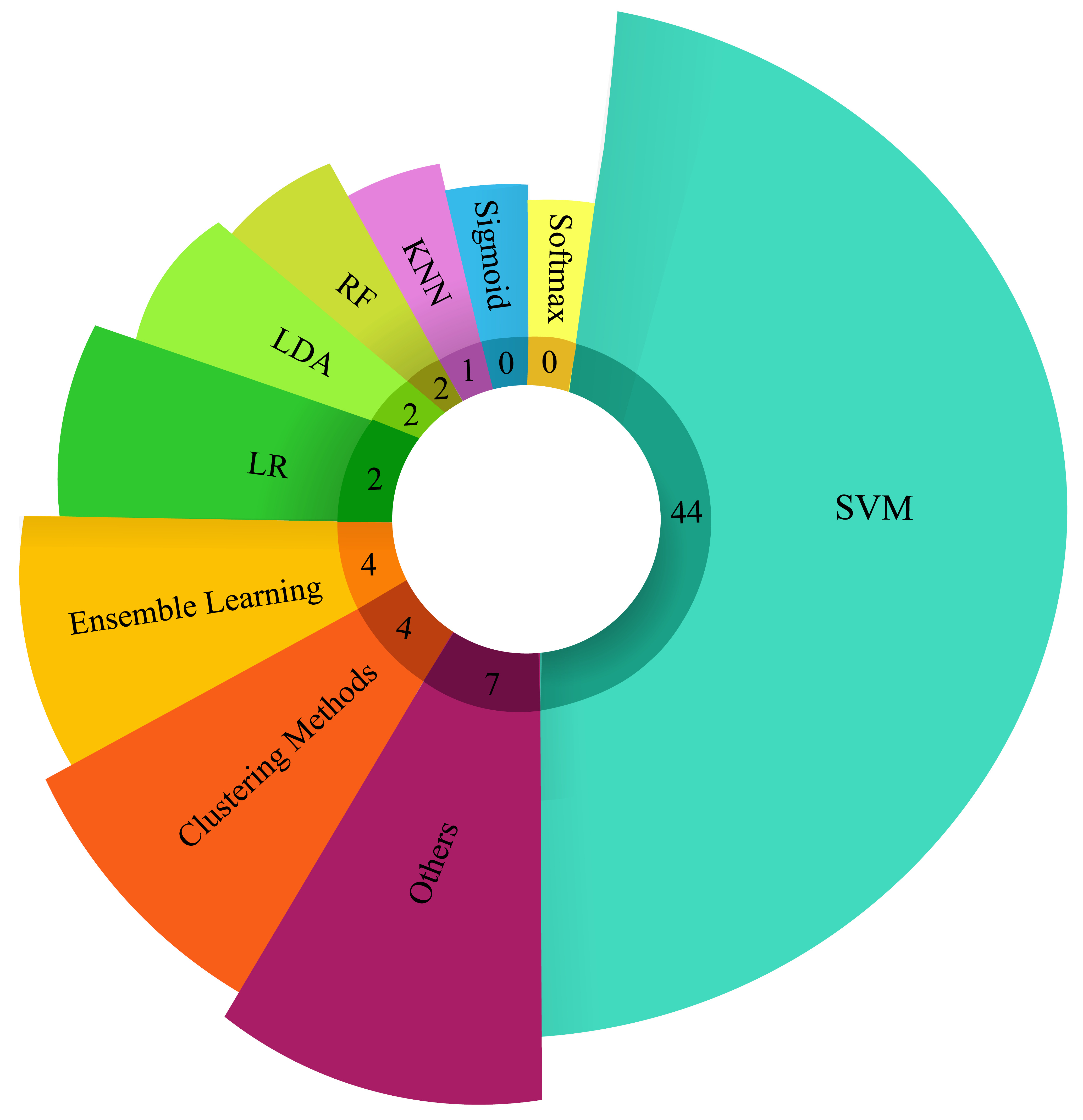}
    \caption{}
    \label{fig:15b}
\end{subfigure}
\caption{Number of MRI-based studies published after 2016 for automated detection of SZ using different AI techniques: (a) DL and (b) ML.}
\label{fig:15}
    
\end{figure}

The number of MRI-based studies published after 2016 for automated detection of SZ using different AI techniques is presented in Figure \ref{fig:15}. It can be noted from Figure \Cref{fig:15a} that, Softmax method is widely used for classification. The SVM classifier is widely used for classification purposes in ML method (Figure \Cref{fig:15b}).

\section{Challenges}

Challenges in the design of automated detection of SZ using MRI modalities and AI techniques are described in this section. Data constraints, algorithmic and hardware problems are the most important challenges in this field, which are discussed below. There are few freely available sfMRI and fMRI datasets (Table \ref{tab:datasetOne}). This has prevented researchers from proposing novel DL models. Hence, more ML models have been proposed which work with a limited number of data samples. Also, the lack of access to all spatially high-resolution sMRI and DTI and spatially-temporally high resolution fMRI datasets are other challenges in this field that avoid researchers to evaluate the effectiveness of simultaneous usage of these MRI modalities to diagnose SZ. The other challenge is to accurately diagnose different types of SZ using sMRI and fMRI modalities. The Schizconnect dataset offers different classes of SZ, but the number of subjects and the variety of chronic disorders are so limited that it is difficult to use them for practical applications. Other available datasets have only SZ and normal classes. Therefore, providing datasets with a large number of subjects and different types of SZ disorders will help researchers to develop a clinically useful system.

In practical applications, we also need to diagnose mental disorders with symptoms close to each other. For example, it is sometimes challenging to diagnose SZ from Bipolar disorder \cite{a255} and ADHD \cite{a256} based on symptoms. Creating a large dataset of patients with these mental disorders, although difficult, can be of great help to physicians in diagnosing SZ accurately.

Early detection and predicting the SZ are very important and challenging tasks. However, due to the difficulty in collecting such datasets (due to the need for longitudinal studies and follow-up of individuals over time), little research has been done in this area and deserves more attention.

Another challenge is related to the use of AI techniques (DL and conventional ML). Implementing CADS based on conventional ML requires a great deal of knowledge in AI. Extracting the distinguishing features which can lead to effective SZ biomarkers is the most important part of CADS. The development of DL architectures for the diagnosis of SZ  has been challenging task due to lack of access to appropriate hardware resources and data availability. Although websites like Google Colab, Amazon, etc. now provide researchers with high computing processors, implementing these methods and using them in the real world still poses many problems.

\section{Conclusion and Future Works}

Schizophrenia is a mental disorder that directly affects the brain, causing symptoms such as abnormal speech and reduced ability to understand. In this work, we have summarized various automated systems developed using MRI neuroimaging modalities to detect SZ early and accurately.  Our findings show that, compared to other diagnostic methods, sMRI and fMRI neuroimaging modalities provide physicians with important information about brain function which helps to accurately diagnose SZ. In these types of neuroimaging modalities, parts of SZ brain do not have a normal structure or function and are usually recognizable. In addition to the benefits of MRI modalities, analyzing this data to diagnose SZ by a physician is complex. To this end, conventional ML and DL techniques have been combined with MRI modalities to assist the clinicians to make an accurate diagnosis of SZ. In this article, a complete review of the diagnosis of SZ with the help of sMRI and fMRI neuroimaging modalities along with DL methods and conventional ML has been done. In the discussion section of this article, a detailed review is conducted on research conducted in the field of DL compared to conventional ML. As discussed, lots of work has been done in automated diagnosis of SZ using conventional ML and DL techniques. The DL networks require a lot of data for training, and the lack of free and available datasets are the main reason for the main challenge in the automated diagnosis of SZ accurately.

Different models of GANs are one of the newest areas of DL that can be used to address this data shortage problem \cite{a121}. In future work, DL networks such as deep convolutional GAN (DCGAN) \cite{a257,a258} will largely address these problems of MRI data shortages to expand DL applications in diagnosing SZ from healthy subjects. Also, as mentioned earlier, the free Schizconnect dataset contains sMRI and fMRI neuroimaging modalities of various schizophrenic disorders. The generation of artificial data from different classes of SZ with the help of GAN architectures can be considered as the future work in designing a CAD system for effective diagnosis of this disease. 
So far, we have reviewed potential future works on the generation of artificial data and increasing the efficiency of CADS for the diagnosis of SZ. Another challenge is the lack of free access to sMRI or fMRI neuroimaging modalities for a particular class of SZ. Zero-shot learning is a new class of AI techniques which can solve the problem of not having access to the data of a class of SZ and is considered as another future work \cite{a259,a260}.

Different types of SZ are growing in less developed countries. Lack of access to specialist physicians to analyze sMRI and fMRI data is always a challenge. In future, the practical implementation of CADS based on DL and cloud computing can greatly provide valuable services to people with these brain disorders. The sMRI or fMRI scan can be sent to the cloud where the accurate DL model can be placed. The result of the model will be sent to the hospital server. After confirmation with the specialist clinician, the diagnosis will results can be sent to the patient. 

\section*{Acknowledgement}
This work was partly supported by the Ministerio de Ciencia e Innovación (España)/ FEDER under the RTI2018-098913-B100 project, and by the Consejería de Economía, Innovación, Ciencia y Empleo (Junta de Andalucía) and FEDER under CV20-45250 and A-TIC-080-UGR18 projects.
\bibliography{mybibfile}

\end{document}